\documentclass{article}
\usepackage{spconf,amsmath,epsfig}
\usepackage{amsfonts}
\usepackage{cite}
\usepackage{algorithmicx}
\usepackage{algorithm}
\usepackage{algpseudocode}
\usepackage{graphicx}
\usepackage{subfigure}
\usepackage{amsmath}
\usepackage{multirow}
\usepackage{array}
\usepackage{xcolor}
\usepackage[colorlinks,
linkcolor=blue,       
anchorcolor=blue,  
citecolor=blue,        
]{hyperref}
\makeatletter
\renewcommand{\@thesubfigure}{\hskip\subfiglabelskip}
\makeatother

\title{Hyperspectral Target Detection Based on Low-Rank Background Subspace Learning and Graph Laplacian Regularization}

\name{Dunbin Shen, Xiaorui Ma, Wenfeng Kong, Jiacheng Tian, Hongyu Wang\thanks{This work was supported by the National Natural Science Foundation of China under Grant U1933104.}}
\address{Faculty of Elec. Info. \& Elec. Engi, Dalian University of Technology, Dalian, China}
\begin{document}
%
\maketitle
\begin{abstract}
Hyperspectral target detection is good at finding dim and small objects based on spectral characteristics. However, existing representation-based methods are hindered by the problem of the unknown background dictionary and insufficient utilization of spatial information. To address these issues, this paper proposes an efficient optimizing approach based on low-rank representation (LRR) and graph Laplacian regularization (GLR). Firstly, to obtain a complete and pure background dictionary, we propose a LRR-based background subspace learning method by jointly mining the low-dimensional structure of all pixels. Secondly, to fully exploit local spatial relationships and capture the underlying geometric structure, a local region-based GLR is employed to estimate the coefficients. Finally, the desired detection map is generated by computing the ratio of representation errors from binary hypothesis testing. The experiments conducted on two benchmark datasets validate the effectiveness and superiority of the approach. For reproduction, the accompanying code is available at \url{https://github.com/shendb2022/LRBSL-GLR}.
\end{abstract}
\section{Introduction}
Due to the integration of spatial and spectral information, hyperspectral images (HSIs) facilitate the detection of dim and small targets, which have a wide range of applications in military and civilian fields\cite{kaul2022hyperspectral}. The objective of hyperspectral target detection (HTD) is to achieve a comprehensive separation between target and background information by prominently highlighting targets while effectively suppressing the background. However, the limited prior knowledge, namely an HSI and a target spectrum, makes this task challenging. 

To achieve this task, various detection methods have been proposed. Euclidean distance and cosine similarity are simple techniques commonly used for direct spectral matching. However, in practical scenarios, their performance is often below expectations due to the inherent diversity and variability of spectra. Statistical assumption-based methods \cite{chang2005orthogonal} operate under the hypothesis that the background follows a multivariate Gaussian distribution, allowing the estimation of pixel probabilities belonging to targets using statistical techniques. However, these methods often experience a decline in performance because the actual distribution of the background is typically complex and does not align with the underlying assumption.  The representation-based methods model HTD as an optimization problem by combining spectral unmixing and some common properties such as sparsity, low rank, and smoothness, which have achieved success in detection performance and physical interpretability. Nevertheless, there are still two problems. First, a background dictionary with completeness and purity is difficult to obtain. Researchers usually use pre-detection\cite{wang2022double}, clustering\cite{xu2015anomaly}, or orthogonal projection\cite{shi2021hyperspectral} methods to select typical background samples to form a background dictionary, in which the atoms still come from pixel samples, carry noise, and may have mixed redundancy. Second, spatial information is rarely employed in most representation-based methods, which limits the detection performance. Some researchers \cite{yang2016hyperspectral, 2021ChengDecomposition} exploit the total variation to promote the spatial smoothness of coefficients but only consider the spatial similarity between a pixel and its four neighbors. 

To tackle the aforementioned issues, an efficient optimizing approach using low-rank representation (LRR) and graph Laplace regularization (GLR) is proposed. To overcome the problem of obtaining a complete and pure background dictionary, we propose a low-rank background subspace learning (LRBSL) method by jointly mining the low-dimensional structure of all pixels. This strategy not only learns the underlying background atoms but also alleviates the contamination of target spectra to them. Moreover,  in order to mine the spatial context features and capture the underlying geometric structure, a local region-based GLR (LRB-GLR) is employed to estimate the coefficients. Combining this regularization and sparse representation, the spatial-spectral features can be effectively extracted for detection. Finally, the desired detection map can be obtained by computing the ratio of representation errors from binary hypothesis testing. The least squares and proposed LRB-GLR are responsible for the hypothesis of modeling background and target, respectively. Experiments are conducted on two benchmark datasets and illustrate the effectiveness of the proposed method in terms of quantitative metrics and visualizations.

\section{Proposed Method}
\subsection{LRBSL}
Given an HSI  ${\bf X} \in \mathbb{R}^{L \times N}$ and a target spectrum ${\bf t} \in \mathbb{R}^{L \times 1}$, where $L$ and $N$ denote the number of bands and pixels, respectively, the linear mixing model is 
\begin{equation}
	{\bf X}= {\bf A}{\bf S}_1 + {\bf t}{\bf S}_2 + {\bf E},
\end{equation}
where ${\bf A} \in \mathbb{R}^{L \times K}$ is the background subspace with $K$ bases, ${\bf S}_1 \in \mathbb{R}^{K \times N}$  and ${\bf S}_2 \in \mathbb{R}^{1 \times N}$ are the encoding coefficients of background and target components, respectively, and ${{\bf E} \in \mathbb{R}^{L \times N}}$ is the representation error. To separate the background and target components, the unknown background subspace needs to be estimated. Here, we propose a two-step optimization method based on sparse representation and LRR. 

In the first step, we fix the joint subspace and estimate the coefficients by incorporating the sparse prior:
\begin{equation}
	\label{model1}
			\min_{{\bf S}} {\frac{1}{2}\|{\bf X} - {\bf B}{\bf S}\|_F^2} + \lambda_1\|{\bf S}\|_1, 
\end{equation}
where $\|\cdot\|_F$ and $\|\cdot\|_1$ denote the Frobenius norm and L1-norm, respectively, ${\bf B}=[{\bf A}, {\bf t}]$ is the joint subspace, ${\bf S}=[{\bf S}_1^T, {\bf S}_2^T]^T$ is the joint coefficients, and $\lambda_1$ is a trade-off parameter. To ensure the independencies of background atoms, ${\bf A}$ is initially generated by truncated singular value decomposition (SVD):
\begin{equation}
	{\bf A},{\bf \Sigma},{\bf U}^T = {\rm svd}({\bf X}, K),
\end{equation}
 where ${\rm svd}$ is the truncated SVD function keeping the top $K$ largest singular values, and ${\bf \Sigma} \in \mathbb{R}^{K \times K}$ and ${\bf U}\in \mathbb{R}^{N \times K}$ are diagonal and semiunitary matrices, respectively.

In the second step, we divide the estimated coefficients ${\bf S}$ into ${\bf S}_1={\bf S}(:K)$ and ${\bf S}_2={\bf S}(K:)$, and employ the low-rank property to learn the background subspace:
\begin{equation}
	\label{model2}
	\min_{{\bf A}} {\frac{1}{2}\|{\bf X} - {\bf A}{\bf S}_1-{\bf t}{\bf S}_2\|_F^2} + \lambda_2\|{\bf A}\|_*,
\end{equation}
where $\|\cdot\|_*$ is the nuclear norm to characterize the low-rank constraint, and $\lambda_2$ is a trade-off parameter. 

Models \eqref{model1} and \eqref{model2} can be efficiently solved using the alternating direction method of multipliers (ADMM). By introducing ${\bf Z}={\bf S}$ and ${\bf D}={\bf A}$, the augmented Lagrangian function can be respectively expressed as
\begin{equation}
	\begin{split}
			{\mathcal{L}}_1 &= \frac{1}{2}\|{\bf X}-{\bf B}{\bf S}\|_F^2 + \lambda_1\|{\bf Z}\|_1 +  \frac{\mu_1}{2}\|{\bf S}-{\bf Z}+\frac{{\bf G}_1}{\mu_1}\|_F^2,\\
		{\mathcal{L}}_2 &= \frac{1}{2}\|{\bf X}-{\bf A}{\bf S}_1 - {\bf t}{\bf S}_2\|_F^2 +\lambda_2\|{\bf D}\|_*  \\&+  \frac{\mu_2}{2}\|{\bf A}-{\bf D}+\frac{{\bf G}_2}{\mu_2}\|_F^2,
	\end{split}
\end{equation}
where ${\bf G}_1$ and ${\bf G}_2$ are two Lagrangian multipliers, and $\mu_1$ and $\mu_2$ are two regularization parameters. The models can be decomposed into the alternated optimization of subproblems. The process of optimization is summarized in Algorithm \ref{alg_Algorithm1}, where $\rm soft$ denotes the soft thresholding function, $\rm svt$ denotes the singular value thresholding function,  and ${\bf I}_j$ is an identity matrix numbered $j$.

\begin{algorithm}[!t]
	\caption{Optimization procedure of LRBSL}
	\begin{algorithmic}[1]
		\State \textbf{Input}:
		$\bf X$, $\bf t$, $\bf B$, $\lambda_1$, $\lambda_2$, $K$.
		\State \textbf{Initialize}:  ${\bf S}^{(0)}={\bf Z}^{(0)}={\bf G}_1^{(0)}={\bf 0}$, ${\bf A}^{(0)}={\bf D}^{(0)}={\bf G}_2^{(0)}={\bf 0}$, $\mu_1=\mu_2=10^{-3}$, $\mu_{max}=10^{10}$, $\gamma=1.2$, $\epsilon=10^{-6}$, $k=0$, $k_{max}=200$.
		\While{$\|{\bf S} - {\bf Z}\|_F \geq \epsilon $ and $k < k_{max}$}
		\State ${\bf Z}^{(k+1)}={\rm soft}({\bf S}^{(k)}+{\bf G}_1^{(k)}/\mu_1, \lambda_1/\mu_1)$;
		\State   
		${\bf S}^{(k+1)}=({\bf B}^T{\bf B}+\mu_1{\bf I}_1)^{-1}$\Statex \qquad \qquad \qquad$({\bf B}^T{\bf X} + \mu_1{\bf Z}^{(k+1)} - {\bf G}_1^{(k)})$;
		\State  ${\bf G}_1^{(k+1)}={\bf G}_1^{(k)}+\mu_1({\bf S}^{(k+1)}-{\bf Z}^{(k+1)})$;
		\State $\mu_1 = {\rm min}(\mu_{max},\mu_1\times\gamma)$;
		\State $k = k + 1$;
		\EndWhile
		\State ${\bf S}_1={\bf S}^{(k)}(:K)$, ${\bf S}_2={\bf S}^{(k)}(K:)$, $k = 0$;
		\While{$\|{\bf A} - {\bf D}\|_F \geq \epsilon $ and $k < k_{max}$}
		\State ${\bf D}^{(k+1)}={\rm svt}({\bf A}^{(k)}+{\bf G}_2^{(k)}/\mu_2, \lambda_2/\mu_2)$;
		\State   
		${\bf A}^{(k+1)}=(({\bf X}-{\bf t}{\bf S}_2){\bf S}_1^T + \mu_2{\bf D}^{(k+1)} - {\bf G}_2^{(k)}) $\Statex \qquad \qquad \qquad$({\bf S}{\bf S}^T+\mu_2{\bf I}_2)^{-1}$;
		\State  ${\bf G}_2^{(k+1)}={\bf G}_2^{(k)}+\mu_2({\bf A}^{(k+1)}-{\bf D}^{(k+1)})$;
		\State $\mu_2 = {\rm min}(\mu_{max},\mu_2\times\gamma)$;
		\State $k = k + 1$;
		\EndWhile
		\State \textbf{Output}:
		${\bf A}^{(k)}$.
	\end{algorithmic}
	\label{alg_Algorithm1}
\end{algorithm}
\subsection{LRB-GLR}	
To depict the spatial similarity between neighboring pixels, a weighted graph $\mathcal{G}=\{{\bf V}, {\bf R}, {\bf W}\}$ is adopted, where ${\bf V}$, ${\bf R}$, and ${\bf W}$ represent the vertex set, edge set, and weight matrix, respectively. To preserve similarity while ignoring dissimilarity, we simply define ${\bf W}$ as the normalized distance using a threshold: ${\bf W}_{i,j}=\left\{
\begin{aligned}
	&1, \;\|x_i-x_j\|_2^2<\sigma,\\
	&0, \;{\rm otherwise}
\end{aligned}
\right.$ where $x_i$ and $x_j$ are two pixels in a given region, and $\sigma$ is the threshold. 

The similarity in the original image space can be transferred to the feature space, so the similarity between two coefficients can be expressed as
\begin{equation}
\sum_{l=1}^{n_l}{{\rm Tr}({\bf S}_l{\bf L}_l{\bf S}_l^T)=}\frac{1}{2}\sum_{l=1}^{n_l}{\sum_{(i,j)\in\Omega_l}{\bf W}_{l_{i,j}}\|s_i-s_j\|_2^2},
\end{equation}
where the whole image is divided into $n_l$ local regions via $\omega \times \omega$ grid for efficient computation, $\Omega_l$ denotes the $l$th region, ${\bf W}_l$ is the weight matrix of  $\Omega_l$, $s_i$ and $s_j$ are the encoding coefficients of $x_i$ and $x_j$ in  $\Omega_l$, ${\rm Tr(\cdot)}$ is the trace of a matrix, and ${\bf L}_l$ is the graph Laplacian matrix of $\Omega_l$, which is calculated by ${\bf L}_l={\bf D}_l-{\bf W}_l$ where ${\bf D}_l$ is a diagonal matrix formed by the sum of each row of ${\bf W}_l$. 

Combining the local spatial similarity and sparse prior of ${\bf S}$, we can get
\begin{equation}
	\label{model4}
	\min_{{\bf S}} {\frac{1}{2}\|{\bf X} - {\bf B}{\bf S}\|_F^2} + \lambda_3\sum_{l=1}^{n_l}{{\rm Tr}({\bf S}_l{\bf L}_l{\bf S}_l^T)}+\lambda_4\|{\bf S}\|_1,
\end{equation}
where the three terms model representation error, local spatial similarity, and sparsity, respectively, and $\lambda_3$ and $\lambda_4$ are two trade-off parameters. 

The model \eqref{model4} can be efficiently solved by ADMM. The companying augmented Lagrangian function is
\begin{equation}
	\begin{split}
		{\mathcal{L}}_3 &= {\frac{1}{2}\|{\bf X} - {\bf B}{\bf S}\|_F^2} + \lambda_3\sum_{l=1}^{n_l}{{\rm Tr}({\bf V}_{1_l}{\bf L}_l{\bf V}_{1_l}^T)}+\lambda_4\|{\bf V}_2\|_1 \\&+{\frac{\mu_3}{2}(\|{\bf S} - {\bf V}_1 + \frac{{\bf H}_1}{\mu_2}\|_F^2 + \|{\bf S} - {\bf V}_2  + \frac{{\bf H}_2}{\mu_2}\|_F^2)},
	\end{split}
\end{equation}
where ${\bf H}_1$ and ${\bf H}_2$ are two Lagrangian multipliers, and $\mu_3$ is a regularization parameter. The solver of the model is summarized in Algorithm \ref{alg_Algorithm2}.
\begin{algorithm}[!t]
	\caption{Optimization procedure of LRB-GLR}
	\begin{algorithmic}[1]
		\State \textbf{Input}:
		$\bf X$, $\bf t$, $\bf A$, $\bf L$, $\lambda_3$, $\lambda_4$.
		\State \textbf{Initialize}:   ${\bf B}=[{\bf A}, {\bf t}]$, ${\bf S}^{(0)}={\bf V}_1^{(0)}={\bf V}_2^{(0)}= {\bf H}_1^{(0)}={\bf H}_2^{(0)}={\bf 0}$, $\mu_3=10^{-3}$, $\mu_{max}=10^{10}$, $\gamma=1.2$, $\epsilon=10^{-6}$, $k=0$, $k_{max}=200$.
		\While{$\|{\bf S} - {\bf V}_1\|_F +\|{\bf S} - {\bf V}_2\|_F  \geq \epsilon $ and $k < k_{max}$}
		 \State ${\bf S}^{(k+1)}=({\bf B}^T{\bf B}+2\mu_3{\bf I}_3)^{-1}$\Statex \qquad \qquad \qquad$({\bf B}^T{\bf X} + \mu_3{\bf V}_1^{(k)} - {\bf H}_1^{(k)} + \mu_3{\bf V}_2^{(k)} - {\bf H}_2^{(k)})$;
		\For{$l=1$ to $n_l$}
		\State ${\bf V}_{1_l}^{(k+1)} = (\mu_3{\bf S}_l^{(k+1)} + {\bf H}_{1_l}^{(k)}) (2\lambda_3{\bf L}_l + \mu_3{\bf I}_4)^{-1}$;
		\EndFor
		\State ${\bf V}_2^{(k+1)}={\rm soft}({\bf S}^{(k+1)}+{\bf H}_2^{(k)}/\mu_3, \lambda_4/\mu_3)$;
		\State  ${\bf H}_1^{(k+1)}={\bf H}_1^{(k)}+\mu_3({\bf S}^{(k+1)}-{\bf V}_1^{(k+1)})$;
		\State  ${\bf H}_2^{(k+1)}={\bf H}_2^{(k)}+\mu_3({\bf S}^{(k+1)}-{\bf V}_2^{(k+1)})$;
		\State $\mu_3 = {\rm min}(\mu_{max},\mu_3\times\gamma)$;
		\State $k = k + 1$;
		\EndWhile
		\State \textbf{Output}:
		${\bf S}^{(k)}$.
	\end{algorithmic}
	\label{alg_Algorithm2}
\end{algorithm}
\subsection{Detection}
For binary hypothesis testing, if the target is absent, the pixel can be represented by the background subspace alone ($H_0$), otherwise by the joint subspace ($H_1$). Therefore, the detection result can be obtained by computing the ratio of representation errors:
\begin{equation}
	d_i = \frac{\|{(\bf X}-{\bf A}{\bf S}_1)_{:,i}\|_2^2}{\|{(\bf X}-{\bf B}{\bf S})_{:,i}\|_2^2},
\end{equation} 
where $d_i$ denotes the detection score of the $i$-th pixel in the image, and $\|\cdot\|_2$ is the L2-norm of vectors. For simplicity, $H_0$ is modeled using least squares and the closed-form solution is ${\bf S}_1=({\bf A}^T{\bf A})^{-1}{\bf A}^T{\bf X}$ while $H_1$ is modeled using LRB-GLR.
\begin{table}[!t]
	\centering
	\caption{Quantitative results and running time (in seconds) of competing methods. Bold highlights the best result while underlined the second.}
	\scriptsize
	\tabcolsep = 4pt
	\renewcommand\arraystretch{1.5}
	\begin{tabular}{llcccccc}
		\hline
		&&SAM&OSP&CSCR&DM-BDL&DSC&Ours \\ 
		\hline
		\multirow{2}*{San Diego I} &AUC&0.9944&0.9964&\underline{0.9986}&0.9954&{\bf 0.9994}&0.9983\\
		\cline{2-8}
		&Time&{\bf 0.01}&\underline{0.07}&21.01&4.83&40.90&5.71\\
		\cline{1-8}
		\multirow{2}*{San Diego II} &AUC&0.9945&0.9821&0.9943&0.9855&\underline{0.9952}&\bf{0.9971}\\
		\cline{2-8}
		&Time&{\bf 0.02}&\underline{0.12}&53.12&2.43&176.69&5.46\\
		\hline
	\end{tabular}
	\label{quality}
\end{table}
\begin{figure}[!t]
	\centering
	\includegraphics[width=0.7\columnwidth]{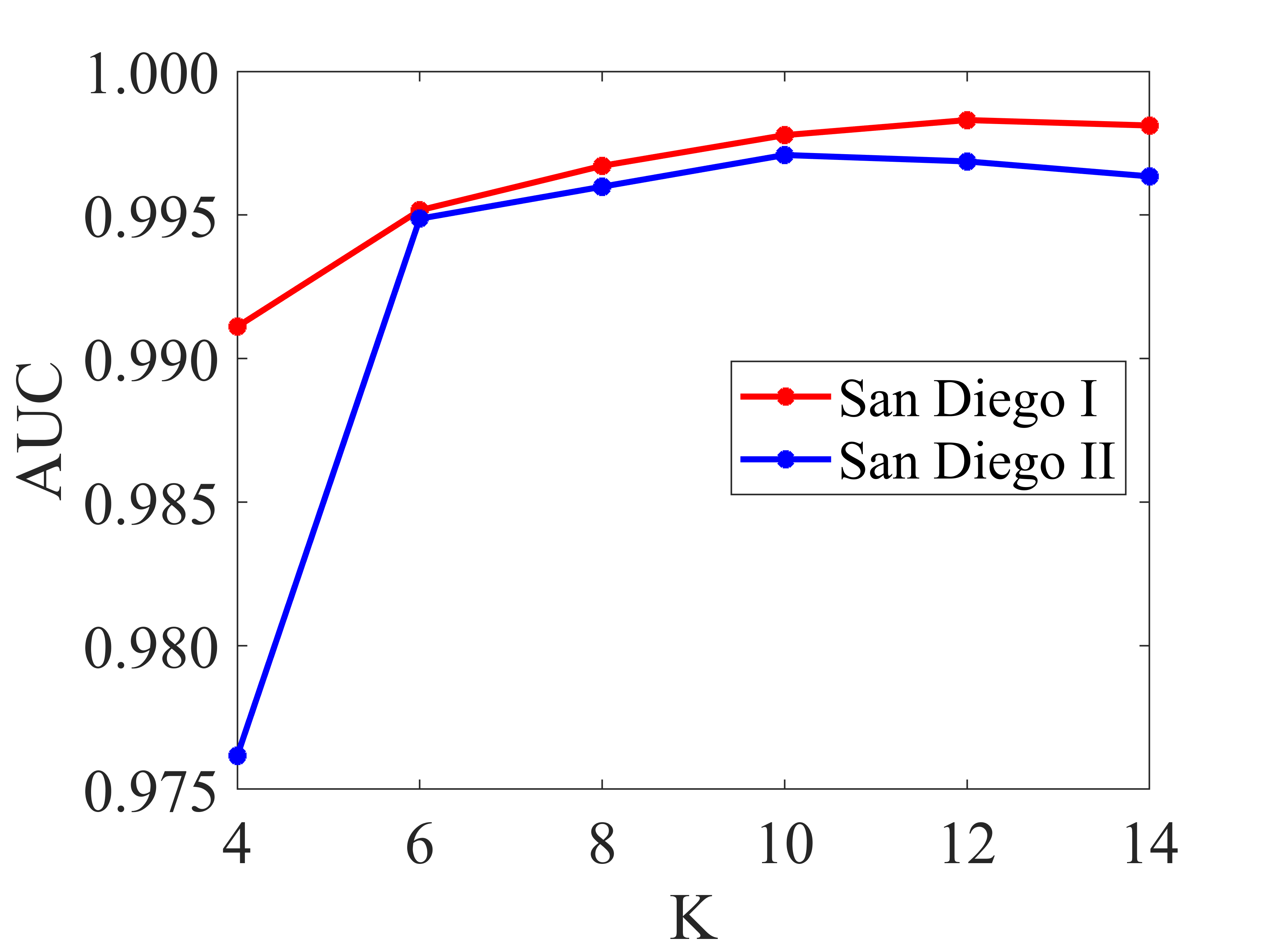}
	\vskip -10pt 
	\caption{Effect of parameter $K$ on the detection performance.}
	\label{parameter_selection}
	\label{sec:con}
\end{figure}
\section{Experiments}
\label{sec:exp}
Two benchmark datasets, San Diego I and San Diego II, are used to evaluate the proposed approach. They were collected using the Airborne Visible/Infrared Imaging Spectrometer and consist of $100\times100$ pixels. After removing water-absorption bands, the datasets contain 189 bands. The targets to be detected are three airplanes, with 58 target pixels and 134 target pixels in each dataset. For simplicity, the average spectrum of all target pixels in each dataset is selected as the target spectrum.

The receiver operating characteristic (ROC) curve and area under the curve (AUC) are used to evaluate the detection performance. In our experiments, we empirically set $\sigma = 0.3$, $\lambda_1=\lambda_2=10^{-4}$, and $\lambda_3=\lambda_4=1$ for the two datasets. By analyzing the impact of parameter $K$ in Fig. \ref{parameter_selection}, we set $K$ as 12 and 10 for the two datasets, respectively.
\begin{figure}[!t]
	\centering
	\subfigure[]{
		\includegraphics[width=0.5\columnwidth]{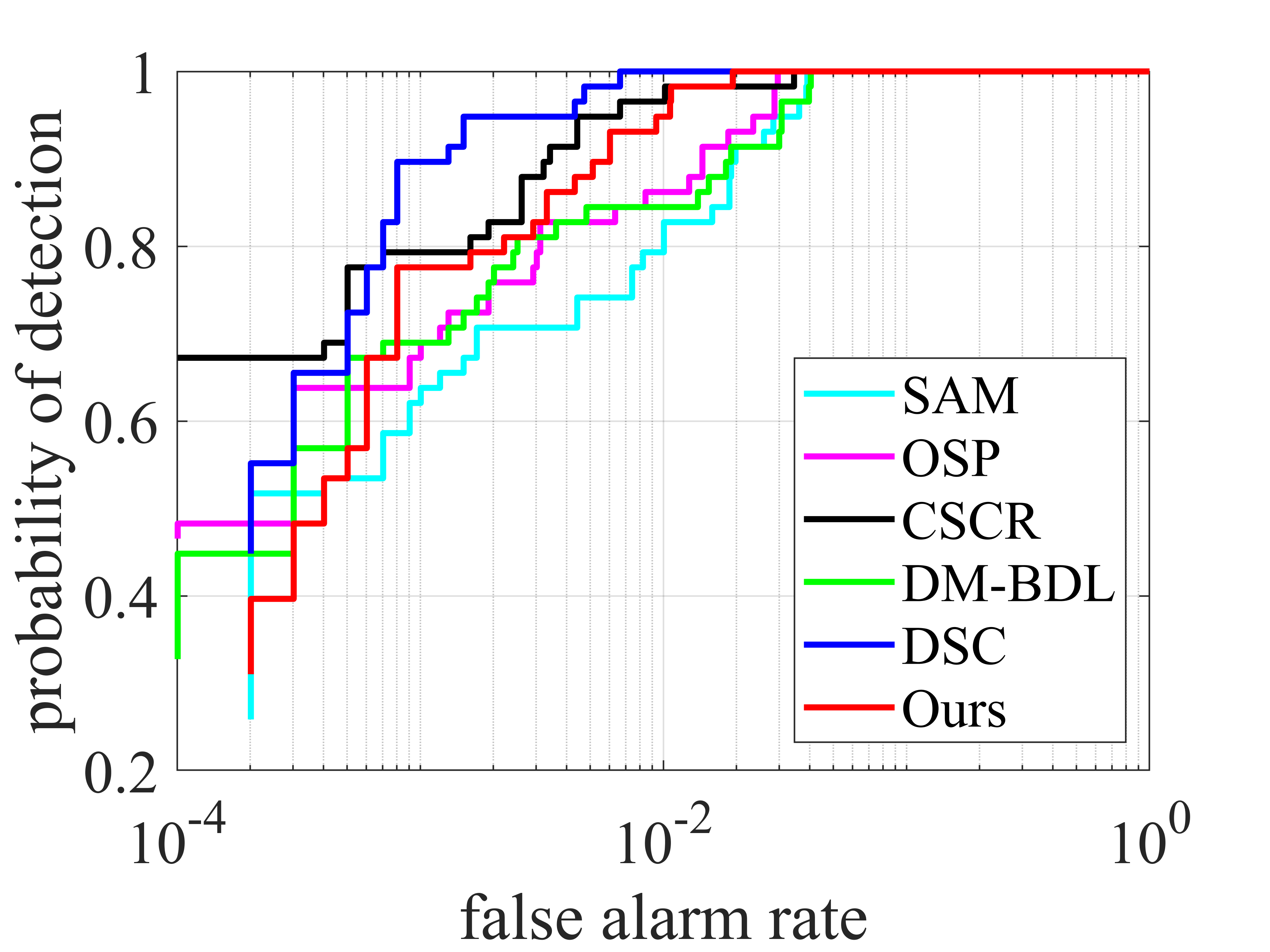}
	}
	\hspace{-15pt}
	\subfigure[]{
		\includegraphics[width=0.5\columnwidth]{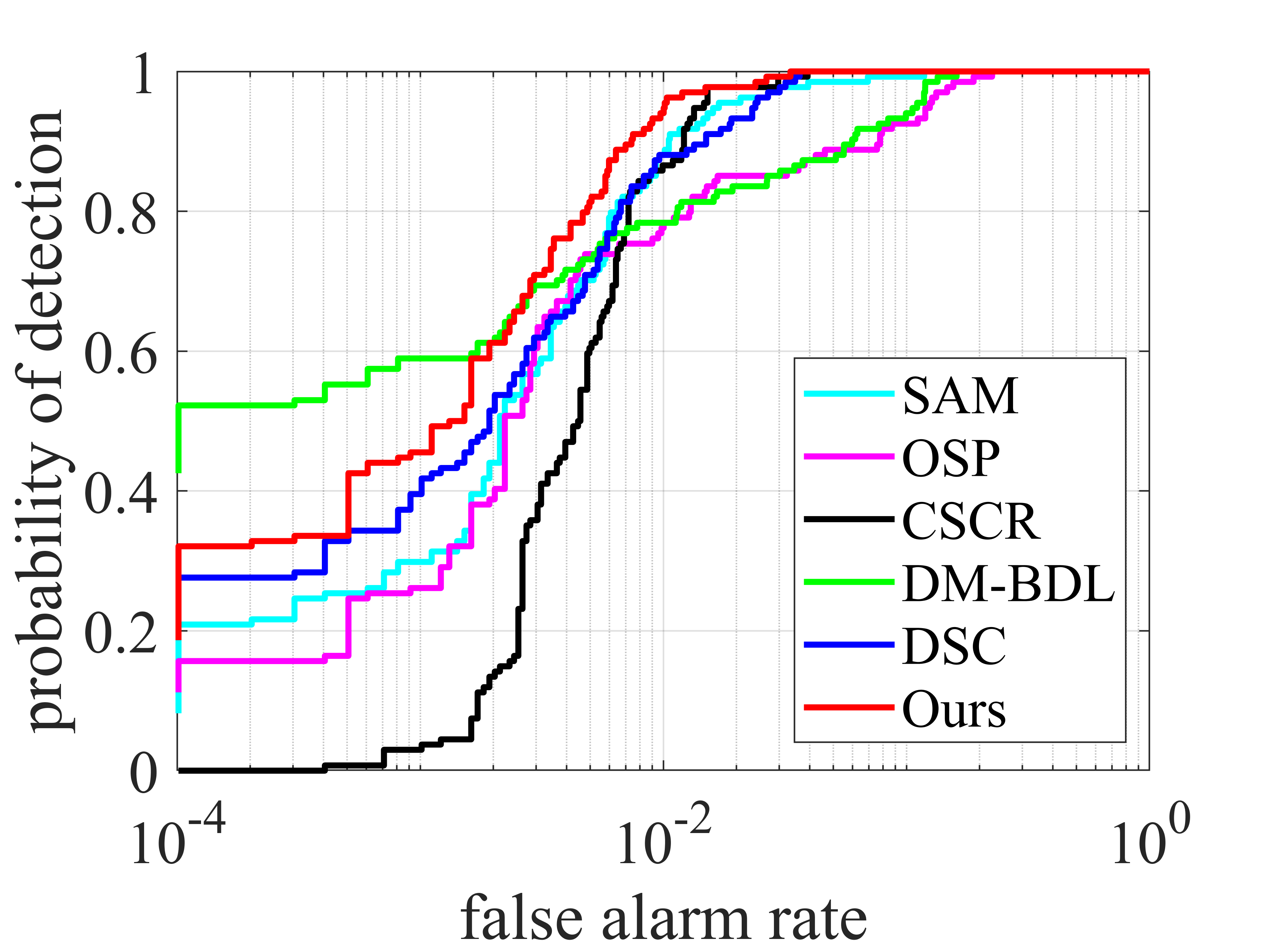}
	}
	\vskip -10pt 
	\caption{ROC curves of competing methods. Left: San Diego I. Right: San Diego II.}
	\label{roc}
\end{figure}
\begin{figure}[!t]
	\centering
	
	\subfigure[]{
		\includegraphics[width=0.1\textwidth]{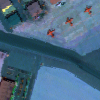}
	}
	\subfigure[]{
		\includegraphics[width=0.1\textwidth]{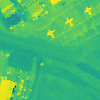}
	}
	\subfigure[]{
		\includegraphics[width=0.1\textwidth]{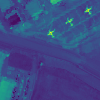}
	} 
	\subfigure[]{
		\includegraphics[width=0.1\textwidth]{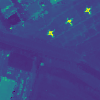}
	}
	\vskip -15pt 
	\subfigure[]{
		\includegraphics[width=0.1\textwidth]{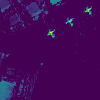}
	}
	\subfigure[]{
		\includegraphics[width=0.1\textwidth]{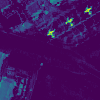}
	}
	\subfigure[]{
		\includegraphics[width=0.1\textwidth]{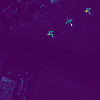}
	}
	\subfigure[]{
		\includegraphics[width=0.1\textwidth]{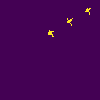}
	}
	\vskip -15pt 
	\subfigure[]{
	\includegraphics[width=0.1\textwidth]{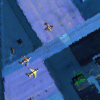}
}
\subfigure[]{
	\includegraphics[width=0.1\textwidth]{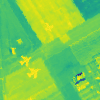}
}
\subfigure[]{
	\includegraphics[width=0.1\textwidth]{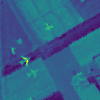}
} 
\subfigure[]{
	\includegraphics[width=0.1\textwidth]{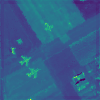}
}
\vskip -15pt 
\subfigure[]{
	\includegraphics[width=0.1\textwidth]{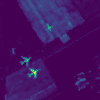}
}
\subfigure[]{
	\includegraphics[width=0.1\textwidth]{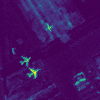}
}
\subfigure[]{
	\includegraphics[width=0.1\textwidth]{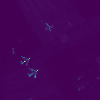}
}
\subfigure[]{
	\includegraphics[width=0.1\textwidth]{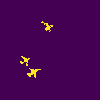}
}
	\vskip -10pt 
	\caption{Detection maps of competing methods. First two rows: San Diego I. Last two rows: San Diego II. From left(top) to right (bottom): false-color image, SAM, OSP, CSCR, DM-BDL, DSC, Ours,  ground truth.} 
	\label{visualized}
\end{figure}

For comparison, five competing methods are selected, including SAM, OSP\cite{chang2005orthogonal}, CSCR \cite{li2015combined}, DM-BDL\cite{2021ChengDecomposition}, and DSC\cite{shen2022dual}. Table \ref{quality} presents the quantitative results, including AUC and time. Combining detection effectiveness and efficiency, it is observed that our method can achieve satisfactory and robust performance on both datasets. In addition,  Figures \ref{roc} and \ref{visualized} display the ROC curves and the final detection maps. It can be found that our method can effectively separate the background and targets and exhibits a promising performance in background suppression.
\section{Conclusion}
This paper presents a fully optimization-based approach for HTD. The proposed approach encompasses a background subspace learning model based on sparse and low-rank representation, enabling the acquisition of a complete and pure background dictionary. Additionally, a model leveraging GLR is introduced to capture the local spatial similarity of coefficients, enabling the employment of spatial-spectral features. By computing representation error ratios through binary hypothesis testing, the desired detection results are obtained. The experimental evaluation of two benchmark datasets validates the effectiveness and superiority of the proposed approach.
\bibliographystyle{IEEEbib}
\bibliography{refs}

\end{document}